\documentclass[11pt]{article}

\usepackage[preprint]{acl}
\usepackage{conference_ACL}
\usepackage{algorithm}
\usepackage[noend]{algpseudocode}
\usepackage{graphicx}
\usepackage{subcaption}
\algrenewcommand{\algorithmicrequire}{\textbf{Input:}}
\algrenewcommand{\algorithmicensure}{\textbf{Output:}}

\title{PHOTON: Hierarchical Autoregressive Modeling for \\ Lightspeed and Memory-Efficient Language Generation}

\author{
 \textbf{Yuma Ichikawa\textsuperscript{1, 2}},
 \textbf{Naoya Takagi\textsuperscript{1}},
 \textbf{Takumi Nakagawa\textsuperscript{1, 3}},
\\
 \textbf{Yuzi Kanazawa\textsuperscript{1}},
 \textbf{Akira Sakai\textsuperscript{1, 4}}
\\
\\
 \textsuperscript{1}Fujitsu Limited,
 \textsuperscript{2}RIKEN Center for AIP,
 \textsuperscript{3}Institute of Science Tokyo,
 \textsuperscript{4}Tokai University
\\
 \small{
   \textbf{Correspondence:} \href{mailto:ichikawa.yuma@fujitsu.com}{ichikawa.yuma@fujitsu.com}
 }
}

\begin{document}
\maketitle

\begin{abstract}
    Transformers operate as \emph{horizontal token-by-token scanners}; at each generation step, attending to an ever-growing sequence of token-level states. This access pattern increases prefill latency and makes long-context decoding more memory-bound, as KV-cache reads and writes dominate inference time over arithmetic operations.
    We propose \underline{\textbf{P}}arallel \underline{\textbf{H}}ierarchical \underline{\textbf{O}}peration for \underline{\textbf{TO}}p-down \underline{\textbf{N}}etworks (\textbf{PHOTON}), a hierarchical autoregressive model that replaces horizontal scanning with vertical, multi-resolution context scanning. PHOTON maintains a hierarchy of latent streams: a bottom-up encoder compresses tokens into low-rate contextual states, while lightweight top-down decoders reconstruct fine-grained token representations in parallel.
    We further introduce \emph{recursive generation} that updates only the coarsest latent stream and eliminates bottom-up re-encoding.
    Experimental results show that PHOTON is superior to competitive Transformer-based language models regarding the throughput-quality trade-off, providing advantages in long-context and multi-query tasks. In particular, this reduces decode-time KV-cache traffic, yielding up to $10^{3}\times$ higher throughput per unit memory.
\end{abstract}

\section{Introduction}\label{sec:introduction}

\begin{figure*}[tb]
    \centering
    \includegraphics[width=\linewidth]{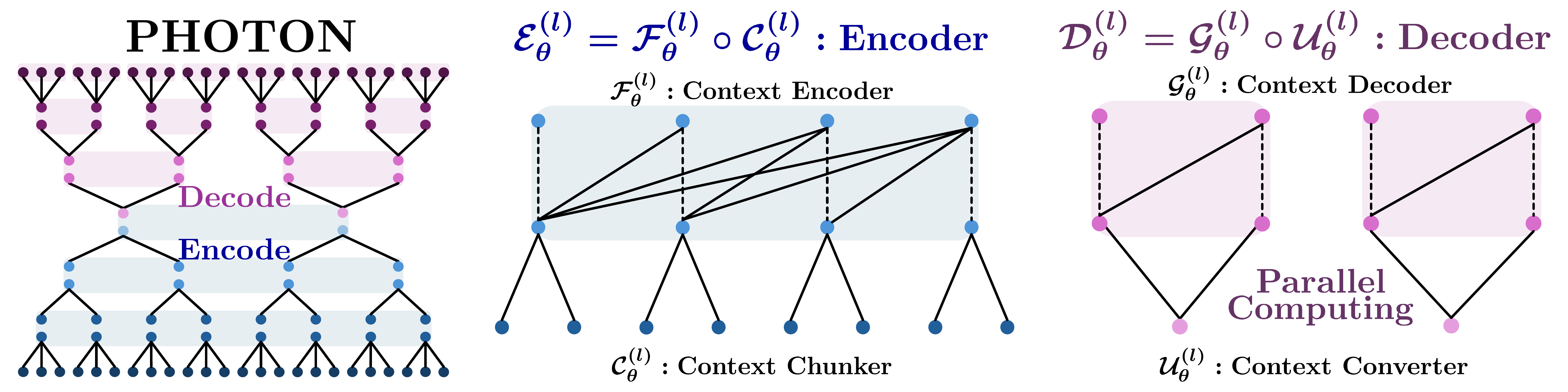}
    \caption{PHOTON overview (left). The hierarchical encoder (middle) compresses token-level states into coarser latent streams via a context chunker and an autoregressive context encoder.
    The hierarchical decoder (right) expands each coarse latent using a context converter and a chunk-local causal decoder with bounded attention.}
    \label{fig:PHOTON}
\end{figure*}

Transformer-based language models have achieved remarkable capabilities; however, the inference cost rapidly increases with context length under recent serving workloads \citep{bahdanau2014neural, vaswani2017attention}.
Even with KV caching, autoregressive Transformers operate as horizontal \emph{token-by-token scanners}; each new token attends to a continually growing flat history of token-level states.
The \emph{prefill} stage computes and stores the KV cache for the entire prompt.
During \emph{decoding}, throughput becomes increasingly memory-bound as the context grows, since each step repeatedly reads from and updates a large KV cache.
As a result, performance is often limited by memory bandwidth rather than computational capacity.
The bottleneck is most pronounced in long-context, multi-query serving.

This raises a simple question:
\textbf{\emph{Must generation remain horizontal token-by-token scanning over a flat history?}}
The structure of natural language suggests otherwise \citep{chomsky2002syntactic,lambek1958mathematics,hauser2002faculty,halle1973prolegomena}.
Natural language is inherently hierarchical: subwords form words, words form sentences, and sentences create documents.
Moreover, coherent generation relies on maintaining an evolving discourse state rather than repeatedly revisiting all fine-grained token representations~\citep{mann1988rhetorical,grosz1986attention,grosz1995centering}.
These observations motivate \emph{vertical scanning}, which represents context through compact coarse states and descends to token-level detail only when necessary.

Hierarchical or multi-scale sequence modeling has recently been explored \citep{pappagari2019hierarchical, han2021transformer, dai2020funnel, nawrot2022hierarchical, nawrot2023dynamic, fleshman2023toucan, mujika2023haed, yu2023megabyte, ho2024block}. 
In particular, Block Transformer~\citep{ho2024block} reduces inference-time KV overhead by separating coarse block-level computation from token-level decoding. However, it employs only a single-level hierarchy that largely serves as a block-structured attention mechanism for efficiency, rather than maintaining a persistent multi-level state that is updated across abstractions during inference.

We introduce \underline{\textbf{P}}arallel \underline{\textbf{H}}ierarchical \underline{\textbf{O}}peration for \underline{\textbf{TO}}p-down \underline{\textbf{N}}etworks (\textbf{PHOTON}), a hierarchical autoregressive model that replaces horizontal token-by-token scanning with multi-resolution vertical scanning over contextual states.
As illustrated in Figure~\ref{fig:PHOTON}, 
PHOTON constructs a hierarchy of latent streams via (i) a bottom-up encoder that compresses tokens into low-rate contextual states and (ii) a top-down decoder stack that progressively reconstructs finer representations using local autoregressive modules with bounded attention.
Since these local decoders operate independently across lower-level contexts, decoding can proceed in parallel across these contexts.
PHOTON is trained using standard next-token prediction along with auxiliary objectives that enforce \emph{recursive consistency} at multiple levels by aligning bottom-up summaries with top-down reconstructions. 
We also propose recursive generation: unlike Block Transformer, PHOTON avoids bottom-up re-encoding of newly generated tokens by directly updating the coarse stream from decoder-side reconstructions.
This design retains only the hierarchical decoder on the GPU during decoding, thus reducing both model residency and KV-cache footprint.

Experiments show that PHOTON achieves a better Pareto frontier in terms of throughput per unit memory and quality compared to both vanilla and Block Transformer baselines. In particular, by reducing decode-time KV-cache traffic, PHOTON achieves up to $10^{3} \times$ higher throughput per unit of memory.


\section{PHOTON}
\subsection{Architecture}
\label{subsec:architecture}

This section presents the architecture of PHOTON, a hierarchical autoregressive language model for sequence modeling across multiple resolutions. PHOTON comprises two components: (i) a hierarchical encoder that progressively compresses the input token sequence into coarser latent streams, and (ii) a hierarchical decoder that reconstructs finer-grained streams in a top-down manner using local autoregressive decoders with strictly bounded attention, enabling parallel decoding across independent lower-level contexts. 
A conceptual overview is provided in Figure \ref{fig:PHOTON}.

\paragraph{Notation.}
Let $\mac{V}$ denote the vocabulary and $t_{1:T}\in\mac{V}^{T}$ be a sequence of length $T$. 
Define $[N]\coloneqq\{1,\ldots,N\}$. 
Let $X^{(0)}\in\mab{R}^{T\times D_0}$ be the embedding matrix, where $D_0$ is the base hidden dimension.
We consider an $L$-level hierarchy indexed by $l\in\{1,\ldots,L\}$. 
Each level $l$ uses a chunk length $C_l\in\mab{N}$ that groups level-$(l-1)$ states into level-$l$ units.
Let $M_0\coloneqq T$ and $M_l\coloneqq \nicefrac{M_{l-1}}{C_l}$ denote the number of units at level $l$. 
Thus, level $l$ consists of $M_l$ contiguous chunks, each spanning $C_l$ level-$(l-1)$ units.
Choose chunk lengths such that $T$ is divisible by $C_{\le L} \coloneqq \prod_{k=1}^{L} C_{k}$, ensuring all $M_{l}$ are integers.
For each $g\in[M_l]$, define the index set of the $g$-th level-$l$ chunk as
\begin{equation}
I_g^{(l)} \coloneqq \{(g-1)C_l + i \mid i\in[C_l]\}\subseteq[M_{l-1}].
\end{equation}
The corresponding level-$(l-1)$ subtensor is defined as follows: 
\begin{equation}
    X^{(l-1)}_{I_g^{(l)}} \coloneqq 
    \ab[X^{(l-1)}_{(g-1)C_l+1},\ldots,X^{(l-1)}_{gC_l}].
\end{equation}
The sets $\{I_g^{(l)}\}_{g=1}^{M_l}$ partition $[M_lC_l]$ into contiguous blocks.
Since $C_l\ge1$, we have $M_l\le M_{l-1}$; thus, $M_0\ge M_1\ge\cdots\ge M_L$. 
All learnable parameters are denoted by $\theta$.

\subsubsection{Hierarchical Encoder}\label{subsec:context-encoder}
At each level $l$, the encoder operates in two stages: (i) it aggregates level-$(l-1)$ representations into chunk-level features, and (ii) it contextualizes these features using an autoregressive context encoder, as shown in Figure \ref{fig:PHOTON}. 
Formally,
\begin{equation}
    X^{(l)} = \mac{E}^{(l)}_{\theta}(X^{(l-1)})
    \coloneqq
    \mac{F}^{(l)}_{\theta}\circ \mac{C}^{(l)}_{\theta}(X^{(l-1)}),
\end{equation}
where the chunker $\mac{C}^{(l)}_{\theta}$ and the context encoder $\mac{F}^{(l)}_{\theta}$ are defined below.

\paragraph{Context Chunker.}
The chunker maps each representation $X_{I_g^{(l)}}^{(l-1)}$ to a single chunk representation:
\begin{equation}
    A^{(l)} \coloneqq [A^{(l)}_{1:M_l}] \in \mab{R}^{M_l\times D_l},
    A_g^{(l)} \coloneqq \mac{C}^{(l)}_{\theta}(X^{(l-1)}_{I_g^{(l)}}).
\end{equation}
In practice, $\mac{C}^{(l)}_{\theta}$ can be implemented by concatenating the representations within a chunk, followed by a linear projection or a one-dimensional convolution. This study utilizes concatenation as a representative instantiation.

\paragraph{Context Encoder.}
The context encoder captures dependencies among chunk embeddings $\{A^{(l)}_g\}_{g=1}^{M_l}$ using an autoregressive Transformer $X^{(l)} = \mac{F}^{(l)}_{\theta}(A^{(l)})$,
producing contextualized chunk-level states at level $l$. Unlike standard Transformers that model sequences token-by-token, our encoder operates over chunk-level contexts, enabling long-range modeling at a coarser temporal resolution.

\subsubsection{Hierarchical Decoder}\label{subsubsec:hierarchical-decoder}
At each level $l$, the decoder (i) converts each level-$l$ latents into a short conditioning prefix and (ii) reconstructs the level-$(l-1)$ stream using a local autoregressive decoder that operates independently within each chunk, as shown in Figure~\ref{fig:PHOTON}.
Formally,
\begin{equation}
    \widehat{X}^{(l-1)} = \mac{D}^{(l)}_{\theta}(\widehat{X}^{(l)})
    \coloneqq
    \mac{G}^{(l)}_{\theta}\circ \mac{U}^{(l)}_{\theta}(\widehat{X}^{(l)}),
\end{equation}
where the converter $\mac{U}^{(l)}_{\theta}$ and the local autoregressive decoder $\mac{G}^{(l)}_{\theta}$ are defined below.

\paragraph{Context Converter.}
To preserve causality, the decoder generates the $g$-th chunk at level $(l-1)$, conditioned on the previous level-$l$ latent.
We introduce a learned starting latent $\widehat{X}^{(l)}_{0}\in\mab{R}^{D_l}$ and define $U^{(l)}_{g-1} \coloneqq \mac{U}^{(l)}_{\theta}(\widehat{X}^{(l)}_{g-1}) \in \mab{R}^{R_l\times D_{l-1}}$ for all $g\in[M_l]$.
In our implementation, $\mac{U}^{(l)}_{\theta}$ is a one-dimensional convolution that expands a single vector into $R_l$ conditioning vectors.

\paragraph{Context Decoder.}
Given $U^{(l)}_{g-1}$, the local decoder generates the level-$(l-1)$ latents autoregressively in each chunk. 
Specifically, for $g \in [M_l]$ and $j \in [C_{l}]$, the lower-level latents are decoded as follows:
\begin{equation}
    \label{eq:local-decoder-ar}
    \widehat{X}_{I_{g}^{(l)}, j}^{(l-1)}
    =
    \mac{G}^{(l)}_{\theta}\ab(
        U^{(l)}_{g-1},
        \widehat{X}_{I_{g}^{(l)}, <j}^{(l-1)};
        \mac{M}_{R_{l},j}^{(l-1)}),
\end{equation}
where the indexed context representations are defined as follows:
\begin{equation}
    \widehat{X}_{I_{g}^{(l)}, j}^{(l-1)} \coloneqq \widehat{X}^{(l-1)}_{(g-1)C_l+j},\widehat{X}_{I_{g}^{(l)}, < j}^{(l-1)} \coloneqq \ab[\widehat{X}_{I_{g}^{(l)}, i}^{(l-1)}]_{i<j}.
\end{equation}
We implement $\mac{G}^{(l)}_{\theta}$ as a causal Transformer applied to the concatenated sequence
$\big[U^{(l)}_{g-1};\, \widehat{X}^{(l-1)}_{I_g^{(l)}, < j}\big]$ using the standard causal mask $\mac{M}^{(l-1)}_{R_l,j} \in \{0,-\infty\}^{(R_l + j - 1)\times(R_l + j - 1)}$.

Under this mask, position $j$ focuses solely on the outputs of the $R_l$ converter and the preceding $(j-1)$ positions within the same chunk.
Thus, the attention span is constrained by $R_l + C_l$ and is independent of the global sequence length $T$.
Since each chunk is conditioned only on its corresponding higher-level context via $U^{(l)}_{g-1}$, the chunks can be decoded in parallel across $g$, resulting in a lightweight local module that enables high-throughput inference.

Overall, PHOTON maps token embeddings through the encoder hierarchy and then back through the decoder hierarchy:
\begin{equation}
\widehat{X}^{(0)}
=
\mac{D}^{(1)}_{\theta} \circ \cdots \circ \mac{D}^{(L)}_{\theta}
\circ
\mac{E}^{(L)}_{\theta} \circ \cdots \circ \mac{E}^{(1)}_{\theta}(X^{(0)}).
\end{equation}
Then, $\widehat{X}^{(0)}$ is projected onto the vocabulary logits.

\begin{takeawaybox}
\textbf{Hierarchical Encoder} compresses the token sequence into \emph{chunk-level latents}, storing long-range context in a compact form.

\vspace{0.50em}

\textbf{Hierarchical Decoder} decodes lower-level latents \emph{in parallel} with local attention, conditioned on higher-level latents.
\end{takeawaybox}

\subsection{Learning Objective}
We train PHOTON using standard next-token prediction with an auxiliary regularizer that promotes consistency among its hierarchical latent streams. Specifically, we minimize the augmented language-modeling loss by including a reconstruction term:
\begin{equation}
    \mac{L}_{\mathrm{PHOTON}}(\theta; \mac{D})
    =
    \mac{L}_{\mathrm{token}}
    + \alpha \mac{L}_{\mathrm{rec}},~~\alpha \in \mab{R}_{+},
\end{equation}
where $\mac{D}$ denotes the training corpus.

\paragraph{Next-Token Loss.}
Our primary objective is autoregressive maximum likelihood. Given the token-level logits computed from $\widehat{X}^{(0)}$, we minimize the negative log-likelihood as follows:
\begin{equation}
    \mac{L}_{\mathrm{token}} = -\sum_{i=1}^{T-1} \log p_{\theta}(t_{i+1}\mid t_{1:i}).
\end{equation}
This preserves PHOTON as a standard decoder-only language model at the output level.

\paragraph{Recursive Loss.}
Next-token supervision alone does not ensure that the model maintains a coherent hierarchy; intermediate latent streams may not be recoverable from the top-down pathway. 
We introduce an auxiliary objective that explicitly enforces hierarchical consistency by matching each encoder state to its corresponding top-down reconstruction at every level:
\begin{equation} 
    \label{eq:recursive-loss} \mac{L}_{\mathrm{rec}} = \sum_{l=1}^{L} \sum_{g=1}^{M_{l}} D(\widehat{X}^{(l-1)}_{I_{g}^{(l)}}, X^{(l-1)}_{I_{g}^{(l)}}),
\end{equation}
where $D(\cdot,\cdot)$ measures the dissimilarity between tensors of the same shape. This study defines $D$ as the cosine distance $(1{-}\mathrm{Cosine Similarity})$, computed for each position and averaged.
By promoting accurate top-down recovery of bottom-up compressed information, this loss improves the fidelity of the multi-rate latent state. As introduced in section \ref{subsec:recursive-generation}, it also fosters recursive consistency, enabling faster inference by allowing decoder-side reconstructions to substitute costly bottom-up re-encoding during recursive generation.


\subsection{Generation}

\subsubsection{Inference Bottlenecks in Transformers}
\label{subsubsection:bottleneck-transformer}

Autoregressive Transformers encounter two inference bottlenecks that become more pronounced with longer contexts. During \emph{prefill}, the model processes the entire prompt in parallel, with latency primarily determined by self-attention and feed-forward computations, making this phase predominantly compute-bound. During \emph{decoding}, generation occurs token by token, repeatedly accessing an expanding KV cache to attend to previous states. Because the KV cache grows linearly with context length and batch size, decoding often becomes memory-bound: throughput is limited by KV cache bandwidth rather than arithmetic throughput. 

\subsubsection{Hierarchical Generation}\label{subsubsec:hierarchical-generation}
PHOTON mitigates sequence-length dependence by transferring global computation to low-rate latent streams and restricting token-level decoding to fixed-size local windows. During inference, each chunk is independently decoded in parallel within the same higher-level context; this process is known as hierarchical generation (HierGen).

\paragraph{Computation.}
At level $l$, the global encoder processes $M_l \approx T/C_{\le l}$ latent units, resulting in a prefill cost that scales as
$\sum_{l=1}^{L}\mac{O}(M_l^2)=\sum_{l=1}^{L}\mac{O}((\nicefrac{T}{C_{\le l}})^2)$,
thereby replacing the vanilla $\mac{O}(T^2)$ dependence with a sum over shorter sequences.
In contrast, each local decoder attends only within its chunk, with a maximum span of $R_l + C_l$, yielding $\mac{O}(1)$ computations per generated token concerning $T$ and facilitating chunk-parallel decoding at a fixed window size.

\paragraph{KV Cache Size.}
The global encoder stores keys and values for $M_l$ latent units at level $l$, resulting in a total global KV storage
of $\sum_{l=1}^{L}\mac{O}(M_l)=\sum_{l=1}^{L}\mac{O}(\nicefrac{T}{C_{\le l}})$.
Each local decoder, however, requires only the KV cache for the \emph{current} chunk, constrained by $\mac{O}(R_l + C_l)$ at level $l$ and is independent of $T$.

\paragraph{KV Cache Access.}
At level $l$, the global context encoder advances once per level-$l$ chunk, i.e., for $M_l$ causal steps over a length-$T$ continuation.
Because step $g$ attends to a prefix of length $\Theta(g)$, the cumulative global KV reads at level $l$ scale as
$\sum_{g=1}^{M_l}\mac{O}(g)=\mac{O}((\nicefrac{T}{C_{\le l}})^2)$,
and the total global KV read traffic is $\sum_{l=1}^{L}\mac{O}((\nicefrac{T}{C_{\le l}})^2)$.
By contrast, each local decoder reads KV only within a bounded window of size $\mac{O}(R_l + C_l)$; over a length-$T$ continuation, this produces total local KV reads
$\mac{O}(\nicefrac{T(R_l + C_l)}{C_{\le(l-1)}})$,
which grows only linearly in $T$.
Therefore, for large $T$, global reads dominate; PHOTON reduces total decode-time KV traffic by shortening the global cached sequences at coarser levels and updating them less frequently, while decoding tokens in parallel across chunks using bounded local attention.

\subsubsection{Recursive Generation}
\label{subsubsec:fast_hier_gen}

\begin{figure}
    \centering
    \includegraphics[width=\linewidth]{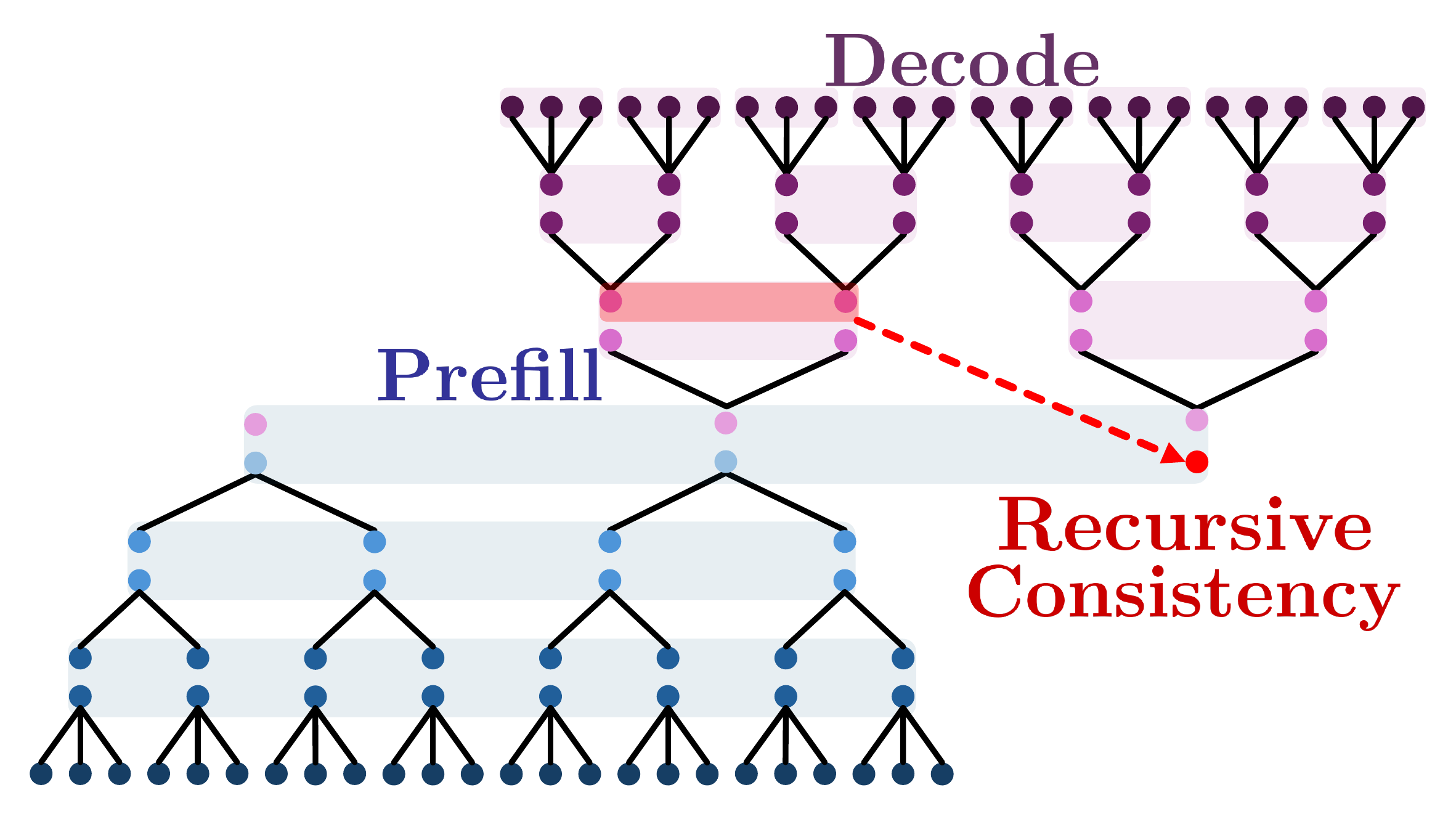}
    \caption{After one-time prefill, recursive generation keeps only the top-level KV cache, decodes meta-contexts top-down, and updates the top-level state via bottleneck summaries.}
    \label{fig:recursive_generation}
\end{figure}

Hierarchical generation confines token-level autoregression to bounded local windows; however, an encoder-consistent implementation typically requires re-encoding newly generated tokens in a bottom-up manner to refresh the hierarchical states on the encoder-side. This re-encoding triggers additional global-attention updates and increases KV-cache traffic across multiple levels.
We introduce recursive generation (RecGen), a decoding schedule that eliminates the need for bottom-up re-encoding. During decoding, only the coarsest stream grows as a global state; it is updated using a summary computed directly from the decoder-side bottleneck reconstruction.  
We refer to the $C_{\le L}$ tokens generated by a single top-level step as a meta-context.

\paragraph{Recursive Decoding.}
Let $\mathrm{KV}^{(L)}_{\le g}$ denote the KV cache of the top-level causal context encoder $\mac{F}^{(L)}_{\theta}$ after processing $g$ top-level inputs, and define its streaming update as
\begin{equation}
    \label{eq:step_top_level}
    \ab(X^{(L)}_{g},\mathrm{KV}^{(L)}_{\le g})
    = \mathsf{STEP}_{\mac{F}^{(L)}_{\theta}} \ab(A^{(L)}_{g},\mathrm{KV}^{(L)}_{\le g-1}).
\end{equation}
After a one-time hierarchical prefill on the prompt, we retain $\mathrm{KV}^{(L)}$ and discard encoder-side caches at levels $1,\ldots,L-1$.
At each meta context step $g$ to $g+1$, we (i) run the top-down decoders conditioned on $X^{(L)}_{g}$ to sample the next tokens and obtain the bottleneck reconstruction $\widehat{X}^{(L-1)}_{I^{(L)}_{g+1}}$, (ii) summarize it using the same chunker 
$\widehat{A}^{(L)}_{g+1}\coloneqq \mac{C}^{(L)}_{\theta}(\widehat{X}^{(L-1)}_{I^{(L)}_{g+1}})$, and (iii) update the coarsest stream by
\begin{equation}
    \label{eq:fast_top_update}
    \ab(X^{(L)}_{g+1},\mathrm{KV}^{(L)}_{\le g+1})
    =
    \mathsf{STEP}_{\mac{F}^{(L)}_{\theta}} \ab(\widehat{A}^{(L)}_{g+1},\mathrm{KV}^{(L)}_{\le g}).
\end{equation}
Therefore, $\mathrm{KV}^{(L)}$ is the only context-growing global cache during decoding; all KV caches in the top-down decoders are local, constrained by fixed windows, and can be discarded at chunk boundaries.

\paragraph{Recursive Consistency and Equivalence.}
RecGen replaces encoder-side re-encoding with a summary computed from the decoder-side bottleneck reconstruction.
A sufficient condition for exact equivalence to HierGen with bottom-up re-encoding is bottleneck \emph{recursive consistency}, namely
$\widehat{X}^{(L-1)} = X^{(L-1)}$.
Under this condition, we have $\widehat{A}^{(L)}_{g}=A^{(L)}_{g}$, meaning both procedures apply the same top-level updates and therefore induce the same distribution over output token sequences.
Equation~\eqref{eq:recursive-loss} explicitly encourages this consistency, making decoder-derived summaries a reliable substitute for encoder-side re-encoding.
We provide the derivation and additional theoretical results in Appendix~\ref{sec:appendix_recursive_generation}.

\paragraph{Inference Cost.}
The prefill remains unchanged; we run the hierarchical encoder once on the prompt. All differences arise during decoding.
In HierGen, the model maintains and advances global encoder streams across all levels, resulting in cumulative KV reads
$\sum_{l=1}^{L}\mathcal{O}(M_l^2)=\sum_{l=1}^{L}\mathcal{O}((\nicefrac{T}{C_{\le l}})^2)$.
In contrast, RecGen retains only the top-level cache $\mathrm{KV}^{(L)}$ and performs a single global streaming update per meta-context, reducing the global KV footprint from $\sum_{l=1}^{L}\mathcal{O}(\nicefrac{T}{C_{\le l}})$ to $\mathcal{O}(\nicefrac{T}{C_{\le L}})$.
As a result, global KV reads collapse to the top-level term $\mathcal{O}((\nicefrac{T}{C_{\le L}})^2)$.
All KV accesses within the top-down decoders are constrained by fixed attention windows, leading to linear increases with $T$.

\begin{takeawaybox}
\textbf{RecGen} speeds up decoding over \textbf{HierGen} by keeping global KV only at the top level and skipping bottom-up re-encoding, halving the GPU-resident model footprint.

\vspace{0.50em}
\centering
\begin{tabular}{@{}lcc@{}}
\toprule
Memory  & \textbf{HierGen} & \textbf{RecGen} \\
\midrule
Size
& $\sum_{l}\mathcal{O}(\nicefrac{T}{C_{\le l}})$
& $\mathcal{O}(\nicefrac{T}{C_{\le L}})$ \\
 Access
& $\sum_{l}\mathcal{O}((\nicefrac{T}{C_{\le l}})^{2})$
& $\mathcal{O}((\nicefrac{T}{C_{\le L}})^{2})$ \\
\bottomrule
\end{tabular}
\end{takeawaybox}

\section{Experiments}\label{sec:experiments}

\begin{table*}[t]
\centering
\caption{
Inference efficiency and language modeling quality for PHOTON and Vanilla/Block Transformer baselines at matched model sizes.
Memory is measured per sample (GiB), throughput in K tokens/s, and TPM in K tokens/s/GiB.
Quality is evaluated by WikiText perplexity and zero-shot accuracy on HS, SCiQ, and ARCe.
}
\label{tab:main}
\resizebox{\textwidth}{!}{
\begin{tabular}{lcccccccccc}
\toprule
\multirow{2}{*}{Models} &
\multicolumn{2}{c}{TPM} &
\multicolumn{2}{c}{Memory} &
\multicolumn{2}{c}{Throughput} &
\multicolumn{4}{c}{PPL and Zero-shot Accuracy } \\
\cmidrule(lr){2-3}\cmidrule(lr){4-5}\cmidrule(lr){6-7}\cmidrule(lr){8-11}
& PF$\uparrow$ & DE$\uparrow$
& PF$\downarrow$ & DE$\downarrow$
& PF$\uparrow$ & DE$\uparrow$
& WikiText$\downarrow$ & HS$\uparrow$ & SCiQ$\uparrow$ & ARCe$\uparrow$ \\
\midrule

\textbf{Vanilla Transformer} \\
600 M & 3.24 & 7.35 & 0.275 & 0.230 & 0.89 & 1.69 & 22.3793 & 41.24 & 72.30 & 43.94 \\
900 M & 3.99 & 6.61 & 0.298& 0.354 & 1.19 & 2.34 & 20.4102 &44.72 & 76.10 & 47.73  \\
1.2 B & 1.21 & 2.56 & 0.439 & 0.390 & 0.53 & 1.00 & 19.6831 & 45.65 & 81.50 & 49.33 \\
\midrule

\textbf{Block Transformer} \\
600 M & 562.73 & 1528.71 & 0.044 & 0.031 & 24.76 & 47.39 & 27.2478 & 37.01 & 70.30 & 42.63 \\
900 M & 485.74 & 1386.05 & 0.054 & 0.038 & 26.23 & 52.67 & 26.3706 & 37.06 & 71.20 & 43.43 \\
1.2 B & 205.00 & 540.20 & 0.070 & 0.051 & 14.35 & 27.55 & 22.8429 & 41.74 & 74.90 & 45.83 \\
\midrule

\textbf{PHOTON} \\
600 M & 1262.58 & 3062.17 & 0.031 & 0.023 & 39.14 & 70.43 & 29.9055 & 35.49 & 67.70 & 42.97 \\
900 M & 1141.62 & 2797.04 & 0.037 & 0.027 & 42.24 & 75.52 & 26.2325 & 38.24 & 69.00 & 44.32   \\
1.2 B & 543.86 & 1216.67 & 0.044 & 0.036 & 23.93 & 43.80 & 23.7863 & 40.70 & 69.30 & 46.25 \\
\bottomrule
\end{tabular}
}
\end{table*}

\paragraph{Architecture and Training Configurations.}
All models are based on LLaMA Transformer architecture~\citep{touvron2023llama}.
We compare PHOTON to vanilla Transformer and block Transformer~\citep{ho2024block}.
For each PHOTON configuration, we tune the baseline architectures to have approximately the same number of trainable parameters; see the details in Appendix \ref{app-subsec:architecture}.
PHOTON is trained on the Pile~\citep{gao2020pile} with a context length of 2048 for one epoch.
For the Vanilla and Block Transformer baselines, we align the training compute budget with the corresponding PHOTON model in terms of total FLOPs.
All experiments are conducted on NVIDIA DGX H200 GPUs.
Additional implementation details are included in Appendix~\ref{sec:appendix_imp_detail}.

\paragraph{Evaluation.}
We evaluate (i) inference efficiency and (ii) language modeling quality.
Following the empirical protocol of \citet{ho2024block}, we measure the per-sample KV-cache memory footprint and throughput ($K$ tokens/s) under two complementary serving regimes:
\emph{prefill-heavy} (PF), with a long prompt and short continuation (2048 input/128 output), and
\emph{decode-heavy} (DE), with a short prompt and long continuation (128 input/2048 output).
To summarize memory efficiency in multi-query serving, we also report throughput-per-memory (TPM), defined as TPM$=$Throughput/Memory in $K$ tokens/s/GiB.
For quality, we report WikiText perplexity (PPL)~\citep{merity2017pointer} and zero-shot accuracy on HellaSwag (HS)~\citep{zellers2019hellaswag}, SciQ~\citep{Welbl2017CrowdsourcingMC}, and ARC-Easy (ARCe)~\citep{Clark2018ThinkYH}.

\subsection{Main Results}\label{subsec:main-results}

\textbf{Throughput-per-Memory Gains.}
PHOTON primarily enhances inference efficiency by reducing the KV cache footprint and associated memory traffic during serving.
This section adopts a representative two-level hierarchy ($L=2$) with chunk lengths $C_{1}=4$ and $C_{2}=4$; Appendix~\ref{subsec:ablations-chunk-length} varies these chunk lengths and demonstrates that PHOTON maintains the most favorable TPM-quality trade-off across context-length settings.
In the main results, we set the recursive reconstruction weight to $\alpha=0.0$ to isolate gains from the hierarchical architecture and bounded local parallel decoding of PHOTON, which differ from the Block Transformer architecture.
Appendix~\ref{subsec:ablations-recursive-loss} further studies $\alpha$ and shows that a moderate value, around $\alpha \approx 0.3$, maximizes downstream zero-shot accuracy, supporting the effectiveness of the proposed recursive regularization.

Table~\ref{tab:main} presents the KV cache memory and throughput per-sample under PF and DE settings.
Across model scales, PHOTON reduces KV-cache memory while increasing throughput, resulting in substantial gains in TPM.
In the PF regime, PHOTON reduces KV-cache memory per sample by $ 8.9\times$ (600M), $ 8.1\times$ (900M), and $ 10.0\times$ (1.2B), while enhancing throughput by $ 44.0\times$, $ 35.5\times$, and $ 45.2\times$, respectively.
In the DE regime, PHOTON reduces KV memory by  $10.0\times $ (600M),  $13.1\times $ (900M), and  $10.8\times $ (1.2B), and increases throughput by  $41.7\times $,  $32.3\times $, and  $43.8\times $.
As a result, PHOTON achieves substantially higher TPM than a vanilla Transformer in both regimes; for the 1.2B model in DE, TPM increases from 2.56 to 1216.67 K tok/s/GiB, corresponding to a $475\times$ improvement. Overall, these results support our central claim that maintaining a hierarchical state reduces decode-time KV traffic and enables more memory-efficient generation.

\begin{figure*}[tb]
  \centering
  \begin{subfigure}[t]{0.49\textwidth}
    \centering
    \includegraphics[width=\linewidth]{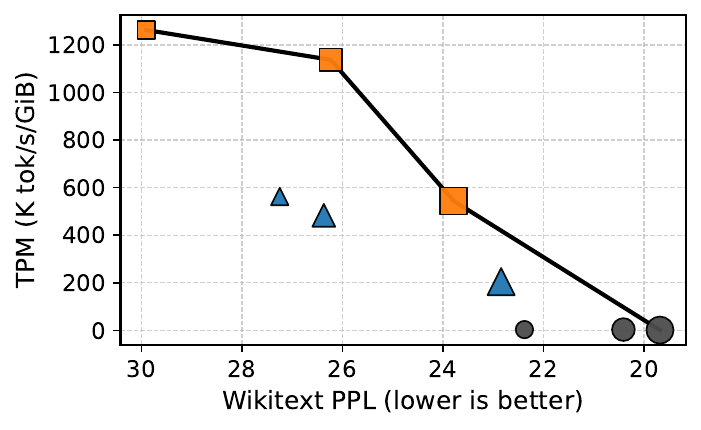}
    \caption{TPM vs.\ WikiText PPL in prefill-heavy (PF).}
    \label{fig:tpm_ppl_pf}
  \end{subfigure}\hfill
  \begin{subfigure}[t]{0.49\textwidth}
    \centering
    \includegraphics[width=\linewidth]{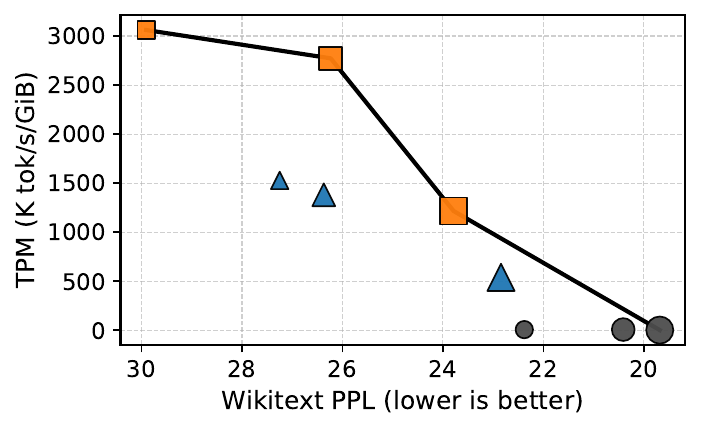}
    \caption{TPM vs.\ WikiText PPL in decode-heavy (DE).}
    \label{fig:tpm_ppl_de}
  \end{subfigure}

  \vspace{0.6em}

  \begin{subfigure}[t]{0.49\textwidth}
    \centering
    \includegraphics[width=\linewidth]{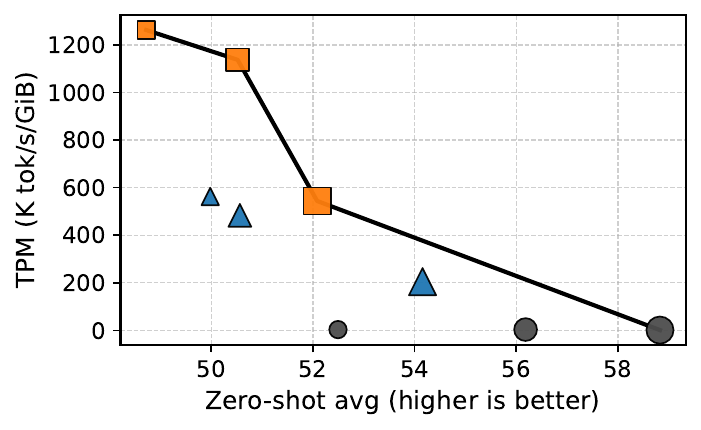}
    \caption{TPM vs.\ zero-shot accuracy (PF)}
    \label{fig:tpm_zs_pf}
  \end{subfigure}\hfill
  \begin{subfigure}[t]{0.49\textwidth}
    \centering
    \includegraphics[width=\linewidth]{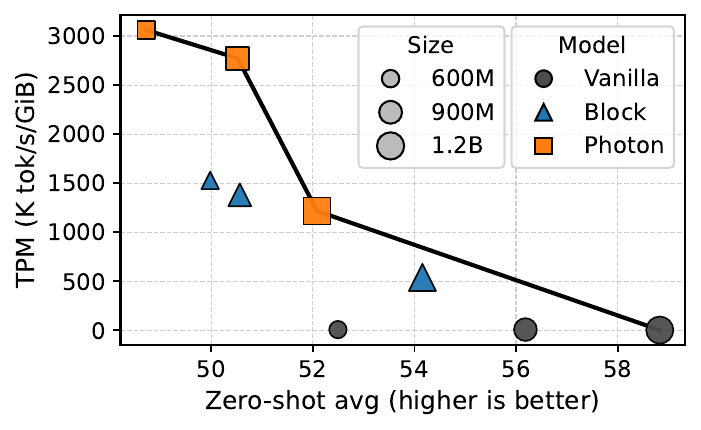}
    \caption{TPM vs.\ zero-shot accuracy (DE)}
    \label{fig:tpm_zs_de}
  \end{subfigure}

  \caption{TPM--quality trade-offs under PF and DE regimes.  Panels (a,b) plot TPM against WikiText perplexity, and panels (c,d) plot TPM against average zero-shot accuracy over HS, SCiQ, and ARCe. The dotted line denotes the Pareto frontier in each panel. Across both regimes and all metrics, PHOTON consistently yields a more favorable TPM--quality frontier than Vanilla and Block Transformers.
  }
  \label{fig:tpm_tradeoff_4panel}
\end{figure*}

\paragraph{TPM vs. Language Modeling Quality.} Figure~\ref{fig:tpm_tradeoff_4panel} shows the trade-off between TPM and language modeling quality.
PHOTON achieves significantly higher TPM with only moderate degradation in WikiText PPL.
 At 600M, TPM increases from 3.24 K tokens/s/GiB to 1262.58 K tokens/s/GiB in the PF regime, and from 7.35 to 3062.17 in DE regime, while PPL slightly increases from 22.38 to 29.91.
For the 1.2B model, TPM rises from 1.21 K tokens/s/GiB to 543.86 K tokens/s/GiB in the PF regime and from 2.56 K tokens/s/GiB to 1216.67 K tokens/s/GiB in DE regime, while PPL moderately increases from 19.68 to 23.79.
As a result, PHOTON achieves significantly higher TPM than Vanilla models in both regimes.
Compared to the Block Transformer, PHOTON achieves higher TPM at both model sizes across both regimes; dominating the Block Transformer concerning the Pareto frontier in Figure~\ref{fig:tpm_tradeoff_4panel}.
This indicates that PHOTON provides a superior TPM-quality operating point, and the saved memory bandwidth can be traded back for quality through test-time scaling within a fixed memory budget.

\subsection{Recursive Generation}\label{subsec:recursive-generation}

\begin{table}[tb]
\centering
\caption{Inference efficiency gains from RecGen over standard HierGen for 600M PHOTON model. Memory usage per sample is reported in GiB, throughput, and TPM under PF and DE settings.}
\label{tab:main_recgen}
\begin{tabular}{llcc}
\toprule
\multirow{2}{*}{Metric} & \multirow{2}{*}{Setting} &
\multicolumn{2}{c}{Generation} \\
\cmidrule(lr){3-4}
& & \textbf{HierGen} & \textbf{RecGen} \\
\midrule
\multirow{2}{*}{TPM} & PF$\uparrow$ & 1262.58 & \textbf{2828.06} \\
                  & DE$\uparrow$ & 3062.17 & \textbf{13642.50} \\
\multirow{2}{*}{Memory} & PF$\downarrow$ & 0.031 & \textbf{0.031} \\
                     & DE$\downarrow$ & 0.023 & \textbf{0.012} \\
\multirow{2}{*}{Throughput} & PF$\uparrow$ & 39.14 & \textbf{87.67} \\
                         & DE$\uparrow$ & 70.43 &  \textbf{163.71} \\
\bottomrule
\end{tabular}
\end{table}

We evaluate RecGen as a drop-in decoding strategy for the 600M PHOTON model under PF and DE settings. We compare RecGen to HierGen, which eliminates the need for bottom-up re-encoding by updating only the coarsest latent stream and utilizing summaries computed from the decoder-side bottleneck reconstruction. 
As shown in Table~\ref{tab:main_recgen}, RecGen enhances efficiency while maintaining nearly unchanged generation behavior. Appendix~\ref{subsec:app-recursive-consistency} supports this observation by demonstrating that the recursive loss steadily decreases throughout training.
In PF regime, RecGen increases throughput from 39.14 to 87.67 K tok/s ($\approx$2.2$\times$) while maintaining the same memory footprint, thus improving TPM by $\approx$2.2$\times$. 
In DE regime, RecGen throughput increased from 70.43 to 163.71 K tok/s ($\approx$2.3$\times$), while memory consumption decreased from 0.023 to 0.012 GiB ($\approx$1.9$\times$), resulting in a 4.5$\times$ increase in TPM.
Notably, since 600M model already achieves roughly 7.35 TPM in Table~\ref{tab:main}, this implies that RecGen can attain a maximum TPM advantage of up to 1,856$\times$ in the corresponding setting.
Overall, RecGen shifts decoding from global KV-cache traffic to bounded local computation, enhancing its suitability for multi-query workloads under fixed GPU memory constraints.

\section{Related Work}\label{sec:related-work}

PHOTON is associated with (i) hierarchical and multi-scale Transformers, (ii) inference efficiency in KV terms, and (iii) global-local modeling in tokenizer-free language models.

\paragraph{Hierarchical and Multi-scale Transformers.}
A substantial body of work reduces the effective length of global attention by introducing intermediate representations.
For example, hierarchical encoders for long documents~\citep{pappagari2019hierarchical}, patch-based or segment-based Transformers~\citep{han2021transformer}, and down-sampled or up-sampled architectures such as Funnel and Hourglass~\citep{dai2020funnel,nawrot2022hierarchical,zhu2021htransformer1d}. These approaches primarily aim for better representations, improved training stability, or reduced training costs. However, during inference, they typically maintain token-level autoregressive decoding, resulting in a KV cache that grows linearly with context length.
In contrast, PHOTON treats hierarchy as an inference primitive; it maintains persistent multi-rate latent streams as global state and confines token-level computation to strictly bounded local refinement.

\paragraph{KV-Efficient Inference.}
Previous work accelerates long-context inference primarily by modifying attention mechanisms, such as sparse or windowed patterns~\citep{child2019sparse,beltagy2020longformer}, and by dynamically retaining a subset of tokens during training or inference~\citep{nawrot2023dynamic,fu2025swat}. Although effective, these methods maintain a single token-level timeline: decoding occurs token by token, which limits inference bandwidth due to repeated KV-cache reads and writes.
PHOTON addresses the same bottleneck but employs a different design principle: it reduces KV traffic through a persistent hierarchical state. Specifically, PHOTON factorizes generation into an encoder-decoder hierarchy that represents global context as low-rate latent streams and produces tokens with strictly bounded local causal decoders. This structure enables parallel decoding of independent chunks conditioned on higher-level latents, thereby shrinking the globally cached sequence and reducing the need for global updates. 

\paragraph{Tokenizer-free Models.}
Tokenizer-free and byte-level language models often adopt global-local hierarchies to model long byte streams tractably, as seen in MEGABYTE~\citep{yu2023megabyte}, SpaceByte~\citep{slagle2024spacebyte}, and learned segmentation approaches~\citep{zakershahrak2025hnetpp,fleshman2023toucan,mujika2023haed}. In these models, the hierarchy induces subword-like units from bytes for a single autoregressive stream: a global module models patches or segments, while a local module reconstructs bytes or characters without maintaining a persistent multi-level state during decoding. PHOTON targets a different objective. Rather than learning subword units from bytes, it introduces higher-level contextual latents to reduce redundant computation and accelerate inference for subword-token language models. This results in a compact encoder-decoder design in which the encoder maintains a persistent coarse context summary, while local decoders use it to refine token-level generation, thereby reducing inference-time memory traffic.

\section{Conclusion}\label{sec:conclusion}
We presented PHOTON, a hierarchical autoregressive language model that replaces horizontal \emph{token-by-token scanning} in Transformers with \emph{vertical} multi-resolution context scanning.
PHOTON builds a hierarchy of latent streams through (i) a bottom-up encoder that compresses the token sequence into low-rate contextual states and (ii) a top-down decoder stack that reconstructs progressively finer representations using local autoregressive modules with bounded attention, enabling parallel decoding across independent contexts.
We further introduced \emph{recursive generation}, updating only the coarsest latent stream during decoding and avoiding bottom-up re-encoding, thus reducing KV-cache growth and memory traffic.
Experiments demonstrate that PHOTON consistently improves the TPM--quality trade-off over strong vanilla and Block Transformer baselines, achieving up to $10^{3}\times$ higher TPM.
These results suggest that the persistent hierarchical state and local reconstruction provide a promising direction for scalable, memory-efficient language generation.

\newpage
\section*{Limitations}

This work has several limitations. First, we train and evaluate PHOTON using a single pretraining corpus and a relatively small set of downstream benchmarks; broader coverage is needed to confirm whether the observed trends generalize across different data mixtures and task families. Second, our largest model contains 1.2 billion parameters, and we have yet to characterize how the efficiency-quality trade-off of PHOTON performs at larger scales. Third, while we report representative results for certain architectural configurations, we do not include a comprehensive sensitivity analysis of key design and training choices, such as chunk sizes and converter widths. A more exhaustive ablation study would help isolate the contributions of each component and clarify which settings are most robust.


\bibliography{ref}

\newpage

\appendix

\section{Additional Theoretical Results}\label{sec:appendix_recursive_generation}

\subsection{Preliminaries}
Fix the number of hierarchy levels $L$ and recall the cumulative meta context length
$B \coloneqq C_{\le L}=\prod_{k=1}^{L}C_k$.
For $g\in\mab{N}$, define the token index set of the $g$-th meta context by
\begin{equation}
    \label{eq:app_metablock_indices}
    J_g \coloneqq \{(g-1)B+1,\dots,gB\}.
\end{equation}
For simplicity, we assume $T$ is divisible by $B$, i.e., $T=GB$ for an integer $G$.
Throughout, we focus on the coarsest causal context encoder $\mac{F}^{(L)}_{\theta}$ and its KV cache.
Let $\mathrm{KV}^{(L)}_{\le g}$ denote the KV cache after processing $g$ top-level inputs.
We express a one-step streaming update by a deterministic operator:
\begin{equation}
    \label{eq:app_step_operator}
    \ab(X^{(L)}_{g},\mathrm{KV}^{(L)}_{\le g})
    =
     \mathsf{STEP}_{\mac{F}^{(L)}_{\theta}} \ab(A^{(L)}_{g},\mathrm{KV}^{(L)}_{\le g-1}),
\end{equation}
where $A^{(L)}_g\in\mab{R}^{D_L}$ is the $g$-th top-level input summary.

\begin{assumption}[Deterministic Streaming Update]
\label{ass:det_step}
    The operator $\mathsf{STEP}_{\mac{F}^{(L)}_{\theta}}$ in Equation~\eqref{eq:app_step_operator} is deterministic given its arguments.
\end{assumption}
Assumption~\ref{ass:det_step} holds for standard causal Transformers with KV caching:
given an input vector and a cache, the next hidden state and the updated cache are uniquely determined.
As an interesting direction for future work, one could relax this assumption by allowing the model to output context \emph{stochastically}, i.e., replacing $\mathsf{STEP}_{\mac{F}^{(L)}_{\theta}}$ with a conditional distribution over updates—potentially enabling richer uncertainty-aware streaming and alternative decoding schedules.

\subsection{Two Decoding Procedures}
We compare two inference procedures that use the same top-down decoder stack
$\{\mac{D}^{(l)}_{\theta}\}_{l=1}^{L}$ and differ only in their construction of the subsequent top-level input.

\begin{definition}[Hierarchical Generation (HierGen)]
\label{def:exact_hier_gen}
Fix a prompt and perform a one-time hierarchical prefill to initialize all encoder-side states and caches.
During decoding, at each meta context step $g\to g+1$, hierarchical generation:
\begin{enumerate}
    \item samples the next meta context tokens $t_{J_{g+1}}$ by running the top-down decoders conditioned on the current top-level state $X^{(L)}_g$;
    \item re-encodes the newly generated tokens bottom-up to obtain the encoder-side bottleneck block
    $X^{(L-1)}_{I^{(L)}_{g+1}}$ and its top-level input summary
    \begin{equation}
    \label{eq:app_A_exact}
    A^{(L)}_{g+1}\coloneqq \mac{C}^{(L)}_{\theta} \big(X^{(L-1)}_{I^{(L)}_{g+1}}\big);
    \end{equation}
    \item advances the top-level stream via Equation~\eqref{eq:app_step_operator} with $A^{(L)}_{g+1}$.
\end{enumerate}
Let $p^{\mathrm{Hier}}_{\theta}$ denote the induced distribution over generated continuations.
\end{definition}

\begin{definition}[Recursive Generation (RecGen)]
\label{def:recursive_gen}
Fix a prompt and perform a one-time hierarchical prefill, retaining \emph{only} the top-level cache $\mathrm{KV}^{(L)}$ thereafter.
During decoding, at each meta context step $g\to g+1$, RecGen:
\begin{enumerate}
    \item samples the next meta context tokens $t_{J_{g+1}}$ by running the same top-down decoders conditioned on $X^{(L)}_g$, and simultaneously obtains the decoder-side bottleneck reconstruction
    $\widehat{X}^{(L-1)}_{I^{(L)}_{g+1}}$ produced by the top-down stack;
    \item forms the next top-level input summary directly from the reconstruction:
    \begin{equation}
    \label{eq:app_A_hat}
    \widehat{A}^{(L)}_{g+1}\coloneqq \mac{C}^{(L)}_{\theta} \big(\widehat{X}^{(L-1)}_{I^{(L)}_{g+1}}\big);
    \end{equation}
    \item advances the top-level stream via Equation~\eqref{eq:app_step_operator} with $\widehat{A}^{(L)}_{g+1}$.
\end{enumerate}
Let $p^{\mathrm{Rec}}_{\theta}$ denote the induced distribution over generated continuations.
\end{definition}

\subsection{Recursive Consistency and Exactness}

\begin{definition}[Recursive consistency]
\label{def:bottleneck_rc}
We say that \emph{recursive consistency} holds if, for every meta context $g\in[G]$,
the reconstruction at the decoder-side bottleneck matches the encoder-side bottleneck state: 
\begin{equation}
    \label{eq:app_bottleneck_rc}
    \widehat{X}^{(L-1)}_{I^{(L)}_{g}} = X^{(L-1)}_{I^{(L)}_{g}}.
\end{equation}
\end{definition}
From Definition~\ref{def:bottleneck_rc}, the reconstruction at the decoder-side bottleneck is equal to the state at the encoder-side bottleneck for each meta-context. Since the top-level input summary is computed deterministically from this bottleneck representation, it follows that the top-level inputs obtained through re-encoding and recursion must coincide, yielding the following Lemma~\ref{lem:input_equality}.
\begin{lemma}
    \label{lem:input_equality}
    Under Definition~\ref{def:bottleneck_rc}, the top-level inputs produced by re-encoding and recursion agree for all $g\in[G]$:
    \begin{equation}
        \label{eq:app_input_equality}
        \widehat{A}^{(L)}_{g} = A^{(L)}_{g}.
    \end{equation}
\end{lemma}

\begin{proof}
By Equation~ \eqref{eq:app_bottleneck_rc} and the definition of the chunker $\mac{C}^{(L)}_{\theta}$,
 \begin{equation}
\widehat{A}^{(L)}_{g}
=
\mac{C}^{(L)}_{\theta} \big(\widehat{X}^{(L-1)}_{I^{(L)}_{g}}\big)
=
\mac{C}^{(L)}_{\theta} \big(X^{(L-1)}_{I^{(L)}_{g}}\big)
=
A^{(L)}_{g},
 \end{equation}
which is Equation~\eqref{eq:app_input_equality}.
\end{proof}
Lemma~\ref{lem:input_equality} demonstrates that recursion yields the same top-level inputs as bottom-up re-encoding at each meta-context step. Combined with Assumption~\ref{ass:det_step}, the resulting top-level hidden states and KV caches evolve identically under recursive and exact hierarchical generation. Consequently, both procedures induce the same conditional distribution at each decoding step, directly implying Theorem~\ref{thm:exactness_recursive_gen}.
\begin{theorem}
    \label{thm:exactness_recursive_gen}
    Assume Assumption~\ref{ass:det_step} and bottleneck recursive consistency (Definition~\ref{def:bottleneck_rc}).
    Then RecGen (Definition~\ref{def:recursive_gen}) induces the same distribution over output token sequences as exact hierarchical generation (Definition~\ref{def:exact_hier_gen}):
     \begin{equation}
    p^{\mathrm{Rec}}_{\theta} = p^{\mathrm{Hier}}_{\theta}.
     \end{equation}
\end{theorem}

\begin{proof}
We couple the two procedures by using the same random draws for token sampling whenever their conditional distributions match.
We prove by induction on $g$ that the top-level state and cache coincide for all $g\in\{0,\dots,G\}$
\begin{equation}
\label{eq:app_state_cache_induction}
    \ab(X^{(L)}_{g},\mathrm{KV}^{(L)}_{\le g})_{\mathrm{Rec}}
    =
    \ab(X^{(L)}_{g},\mathrm{KV}^{(L)}_{\le g})_{\mathrm{Hier}}.
\end{equation}
They coincide at $g=0$ after the shared prompt prefill. Assume Equation~\eqref{eq:app_state_cache_induction} holds at step $g$.
Conditioned on the identical top-level state $X^{(L)}_g$, both procedures run the same top-down decoder stack with the same causal masking, thus defining the same conditional distribution over the next meta context $t_{J_{g+1}}$.
Using the coupling, both procedures sample the same realization of $t_{J_{g+1}}$.

By Lemma~\ref{lem:input_equality}, the top-level inputs then agree:
$\widehat{A}^{(L)}_{g+1}=A^{(L)}_{g+1}$.
Finally, by Assumption~\ref{ass:det_step} and the shared cache $\mathrm{KV}^{(L)}_{\le g}$, the deterministic update Equation~ \eqref{eq:app_step_operator} yields identical
$\big(X^{(L)}_{g+1},\mathrm{KV}^{(L)}_{\le g+1}\big)$.
Thus, Equation~\eqref{eq:app_state_cache_induction} holds for $g+1$.
By induction, the coupled procedures generate identical token sequences almost surely, implying that their induced distributions coincide.
\end{proof}

Theorem~\ref{thm:exactness_recursive_gen} shows that exactness requires equality only at the bottleneck interface $L\to L{-}1$.
No assumptions are needed about lower-level reconstructions to establish equality of the \emph{output} distribution.

\begin{figure}[tb]
    \centering
    \includegraphics[width=\linewidth]{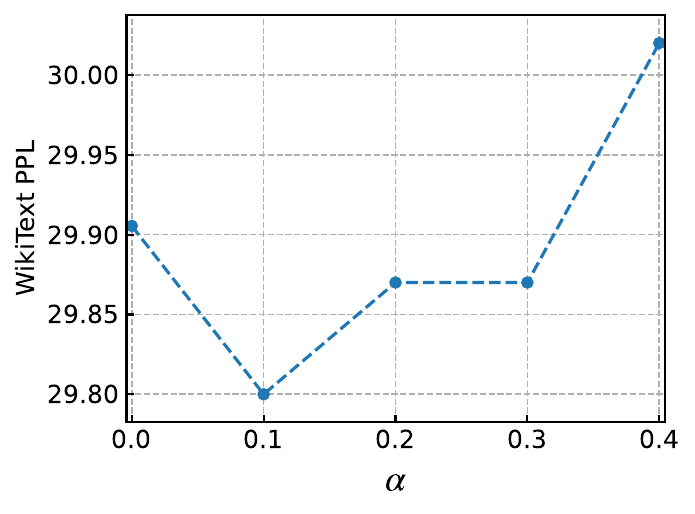}
    \includegraphics[width=\linewidth]{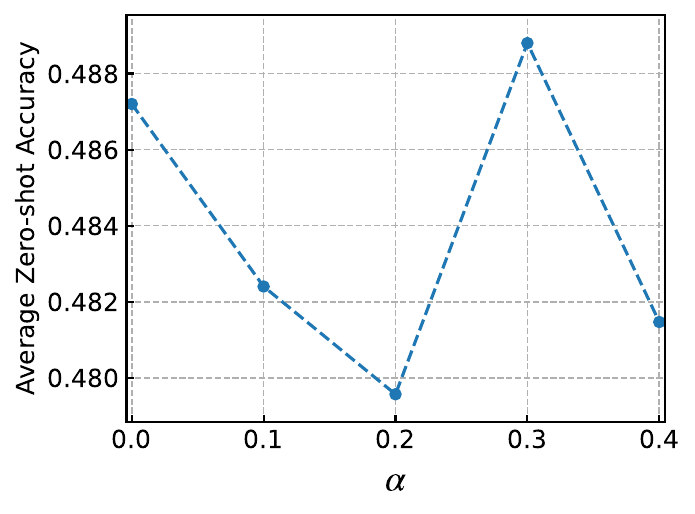}
    \caption{Ablations over $\alpha$. Top: WikiText perplexity (PPL; lower is better). Bottom: Average zero-shot accuracy (higher is better), computed as the mean over ARC-Easy, HellaSwag, and SciQ.}
    \label{fig:ablation_alpha}
\end{figure}

\section{Additional Experiments}\label{app-sec:additional-experiments}

\subsection{Ablations over Strength of Recursive Loss}
\label{subsec:ablations-recursive-loss}
We ablate the hyperparameter $\alpha$ and evaluate its impact while keeping all other settings fixed, including the 4$\times$4 block configuration.
Figure~\ref{fig:ablation_alpha} summarizes the results in terms of WikiText perplexity (PPL) and \emph{Average Zero-shot Accuracy}, with the latter computed as the mean accuracy over ARC-Easy, HellaSwag, and SciQ.
WikiText perplexity remains largely stable for $\alpha \le 0.3$ and increases at $\alpha=0.4$, indicating that an excessively strong recursive loss can negatively impact language modeling. Average zero-shot accuracy exhibits a non-monotonic trend: it decreases from $\alpha=0.0$ to $\alpha=0.2$, then recovers and peaks at $\alpha=0.3$, before declining again at $\alpha=0.4$. These results suggest that a moderate nonzero $\alpha$ can provide a slight downstream gain over $\alpha=0.0$, but the benefit is sensitive to tuning and does not improve monotonically with $\alpha$.

\subsection{Ablations over Chunk Lengths}
\label{subsec:ablations-chunk-length}
\begin{table}[tb]
\centering
\caption{Ablation study of chunk length. Memory usage per sample is reported in GiB, throughput, and TPM under PF and DE settings.}
\label{tab:chunk_length_ablation}
\begin{tabular}{llcc}
\toprule
\multirow{2}{*}{Metric} & \multirow{2}{*}{Setting} &
\multicolumn{2}{c}{Chunk Length $(C_{1}, C_{2})$} \\
\cmidrule(lr){3-4}
& & $(2, 2)$ & $(4, 4)$ \\
\midrule
\multirow{2}{*}{TPM} & PF$\uparrow$ & 175.45 & 1262.58 \\
                  & DE$\uparrow$ & 418.30 & 3062.17 \\
\multirow{2}{*}{Memory} & PF$\downarrow$ & 0.066 & 0.031 \\
                     & DE$\downarrow$ & 0.053 & 0.023 \\
\multirow{2}{*}{Throughput} & PF$\uparrow$ & 11.58 & 39.14 \\
                         & DE$\uparrow$ & 22.17 & 70.43 \\
PPL $\downarrow$ & - &24.28& 29.91 \\
Accuracy$\uparrow$ & - & 52.01 & 48.72  \\
\bottomrule
\end{tabular}
\end{table}

Chunk lengths $\{C_1,C_2\}$ directly determine the compression ratio of PHOTON's hierarchical state and, the
\emph{effective context length} processed by the global (coarse) encoders. With a two-level hierarchy ($L=2$) and
a fixed token length $T$, the top-level sequence length is $M_2 = T/(C_1C_2)$: smaller chunks yield longer latent
streams and more frequent global updates, which can improve modeling fidelity while increasing global KV-cache traffic.
To quantify this trade-off, we adhere to the primary setup outlined in Section~\ref{sec:experiments}, keeping all hyperparameters
fixed (600M model, Pile pretraining, training context length of 2048, and the same evaluation protocol). We compare a
finer hierarchy ($C_1{=}2,C_2{=}2$; 2$\times$2) to the default configuration ($C_1{=}4,C_2{=}4$; 4$\times$4).

Table~\ref{tab:chunk_length_ablation} shows that 2$\times$2 substantially improves quality, reducing WikiText PPL
from 29.91 to 24.28 and increasing average zero-shot accuracy from 48.72 to 52.01, while incurring the expected cost in efficiency:
under PF, throughput decreases from 39.14 to 11.58 K tok/s, and KV memory rises from 0.031 to 0.066 GiB (TPM drops
from 1262.58 to 175.45); under DE, throughput decreases from 70.43 to 22.17 K tok/s, and memory increases from
0.023 to 0.053 GiB (TPM drops from 3062.17 to 418.30). These results highlight $\{C_1,C_2\}$ as a simple lever for navigating the TPM--quality Pareto frontier: smaller chunks mitigate the compression bottleneck and increase the
temporal resolution of the global latent stream, thereby improving perplexity and downstream accuracy, while larger chunks
maximize throughput by minimizing global-state growth and KV traffic. Notably, even the quality-oriented 2$\times$2
configuration remains significantly more memory-efficient than a vanilla Transformer at the same scale, as shown in Table~\ref{tab:main},
while also narrowing the quality gap. This suggests that PHOTON can flexibly convert some of its efficiency gains back into modeling performance without deviating from the Pareto frontier.

\subsection{Recursive Consistency}
\label{subsec:app-recursive-consistency}

\begin{figure}[tb]
  \centering

  \begin{subfigure}[t]{0.95\columnwidth}
    \centering
    \includegraphics[width=\linewidth]{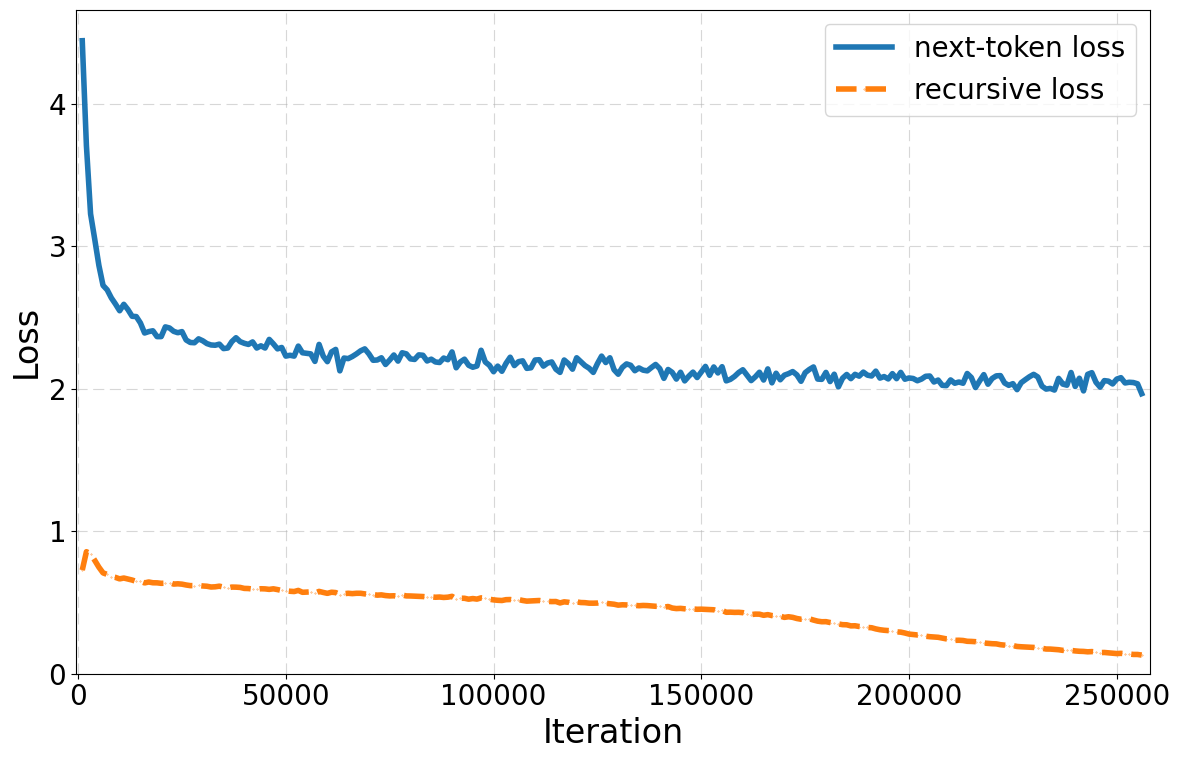}
    \caption{\textbf{Next-token loss vs.\ recursive loss.}
    The recursive loss decreases steadily during training and becomes small in absolute magnitude. Although it is not directly comparable to the token-level decoding loss due to different units, its convergence to a low value indicates a small bottleneck mismatch.}
    \label{fig:token_vs_recursive_loss}
  \end{subfigure}

  \vspace{0.6em}

  \begin{subfigure}[t]{0.95\columnwidth}
    \centering
    \includegraphics[width=\linewidth]{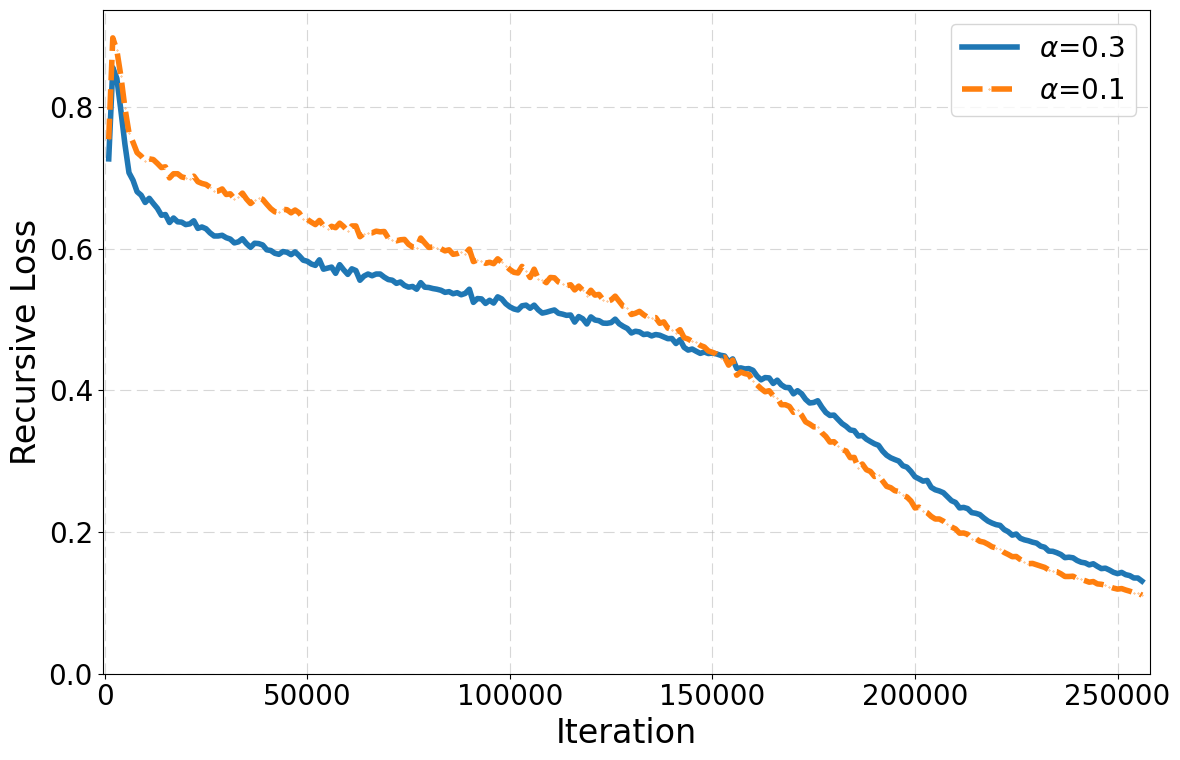}
    \caption{\textbf{Effect of $\alpha$.}
    Recursive-loss trajectories for two values of $\alpha$ show that a smaller weight achieves a slightly lower loss late in training. In both cases, the loss continues to decrease, suggesting that extended optimization can further strengthen recursive consistency.}
    \label{fig:recursive_loss_alpha}
  \end{subfigure}  \label{fig:recursive_consistency_dynamics}
\end{figure}

\paragraph{Recursive Loss vs. Next-Token Loss.}
RecGen replaces bottom-up re-encoding with summaries computed from the decoder-side bottleneck reconstruction, making \emph{recursive consistency} between encoder states and top-down reconstructions critical. A practical indicator of this consistency is the recursive loss in Eq.~\ref{eq:recursive-loss}. Figure~\ref{fig:token_vs_recursive_loss} reports training dynamics for the 600M PHOTON model with $\alpha=0.3$ ($(C_{1}, C_{2})=(4, 4)$): the recursive loss steadily decreases and becomes small in absolute value by the end of training. Although the token-level negative log-likelihood and the cosine-distance-based recursive loss are not directly comparable in scale or units, the diminishing recursive term suggests that the top-down pathway can closely approximate the encoder-side bottleneck state. This provides empirical support that RecGen should only minimally perturb generation behavior when the bottleneck mismatch is sufficiently small. We also find that the recursive loss decreases even when $\alpha=0$, indicating that the hierarchical architecture itself may encourage self-consistent summaries; a more detailed characterization of how residual mismatch translates into long-horizon distributional drift under RecGen is left for future work

\paragraph{Evaluating Likelihood of RecGen.}
RecGen updates the global state using model-generated reconstructions instead of ground-truth tokens, making exact likelihood evaluation under the \emph{RecGen-induced} process difficult to incorporate into standard teacher-forcing pipelines. As a result, computing perplexity and zero-shot accuracy \emph{under RecGen} in a fully consistent likelihood-based protocol is non-trivial. We therefore report perplexity and zero-shot results under standard teacher-forced execution, i.e., HierGen, and leave a more rigorous likelihood-based evaluation of RecGen for future work.

\paragraph{Dependence on  Reconstruction Weight $\alpha$.}
Figure~\ref{fig:recursive_loss_alpha} compares the recursive-loss trajectories of the 600M PHOTON model with $\alpha \in \{0.3, 0.1\}$. A smaller weight results in a slightly lower recursive loss late in training, and in both cases, the loss continues to decrease until the end of optimization. These trends suggest that recursive consistency can be improved not only by tuning the auxiliary objective but also by extending the training duration, which may further narrow the gap between RecGen and the encoder-consistent decoding procedure.

\section{Additional Implementation Details}
\label{sec:appendix_imp_detail}

\subsection{Training Setting}\label{subsec:training-setting}
We employ the 134B Pile-uncopyrighted dataset, consisting of $177{,}008{,}913$ documents and $134{,}217{,}728{,}000$ tokens~\cite{gao2020pile}. 
We also utilize the Llama tokenizer, which has a vocabulary size of $\#(\mac{V}) = 32{,}000$. 
For training, we set the total batch size to 256, the context window to 2048, and the number of training epochs to 1 for PHOTON. 
For both Vanilla Transformer and Block Transformer, we align the training compute budget with the corresponding PHOTON model in terms of total Flops. 
We employ the Adam optimizer with a learning rate of $3 \times 10^{-4}$ and a warm-up period of 3,000 steps. 
The scalar hyperparameters of PHOTON are set to $\alpha = 0.0$.
We further ablate the strength of the recursive loss by sweeping $\alpha$, see Appendix~\ref{subsec:ablations-recursive-loss}; increasing $\alpha$ generally improves downstream zero-shot performance.
All experiments are conducted on an NVIDIA DGX H200 system.

\subsection{Architecture}\label{app-subsec:architecture}
The vanilla Transformer employs an LLaMA architecture with adjusted parameter sizes. The configurations for the 600M, 900M, and 1.2B models are presented in Table~\ref{tab:pv600M}, Table~\ref{tab:pv900M}, and Table~\ref{tab:pv1.2B}, respectively.
Block Transformer follows the architecture proposed in the original paper with adjustment of parameters. The specifications for the 600M, 900M and 1.2B models are provided in Table~\ref{tab:pb600M},  Table~\ref{tab:pb900M}, and Table~\ref{tab:pb1.2B}, respectively.
The block decoders and token decoders in PHOTON are based on the LLaMA decoder architecture. The parameter configurations for the 600M, 900M, and 1.2B models are presented in Table~\ref{tab:pp600M},  Table~\ref{tab:pp900M}, and Table~\ref{tab:pp1.2B}, respectively.


\begin{table*}[tb]
\centering
\caption{Parameter breakdown for the Vanilla Transformer (600M).}
\label{tab:pv600M}
\begin{tabular}{lrrr}
\toprule
Module & Hidden/ Int. / Layers & Params   \\
\midrule
Token Embedding (vocab=32000, d=1664)    & 1664 /-- / -- & 53,248,000  \\
Transformer Blocks (atten h=32, key value h=32, head d =52)   & 1664/4096/16  & 504,418,304  \\
Final Norm (RMSNorm)                 & 1664/ -- / -- & 1,664   \\
LM Head                               & 1664/ -- / -- & 53,248,000   \\

\midrule
Total                                 & --       & 610,915,968  \\
\bottomrule
\end{tabular}
\end{table*}
\begin{table*}[tb]
\centering
\caption{Parameter breakdown for the Vanilla Transformer (900M).}
\label{tab:pv900M}
\begin{tabular}{lrrr}
\toprule
Module & Hidden / Int. / Layers & Params  \\
\midrule
Token Embedding (vocab=32000, d=1792) & 1792 / -- / -- & 57,344,000  \\
Transformer Blocks (atten h=32, key value h=32, head d =56)  & 1792/4608/20  & 752,424,960  \\
Final Norm (RMSNorm)                 & 1792/ -- / -- & 1792   \\
LM Head                               & 1792/ -- / -- & 57,344,000  \\
\midrule
Total                                 & --       & 867,114,752  \\
\bottomrule
\end{tabular}
\end{table*}
\begin{table*}[tb]
\centering
\caption{Parameter breakdown for the Vanilla Transformer (1.2B).}
\label{tab:pv1.2B}
\begin{tabular}{lrrr}
\toprule
Module & Hidden / Int. / Layers & Params  \\
\midrule
Token Embedding (vocab=32000, d=1920) & 1920 / -- / -- & 61,440,000  \\
Transformer Blocks (atten h=32, key value h=32, head d =60)  & 1920/5120/24  &  1,061,775,360  \\
Final Norm (RMSNorm)                 & 1920/ -- / -- & 1,920   \\
LM Head                               & 1920/ -- / -- & 61,440,000  \\
\midrule
Total                                 & --       & 1,184,657,280  \\
\bottomrule
\end{tabular}
\end{table*}
\begin{table*}[tb]
\centering
\caption{Parameter breakdown for the Block Transformer (600M).}
\label{tab:pb600M}
\begin{tabular}{lrrr}
\toprule
Module & Hidden / Int. / Layers & Params  \\
\midrule
Embedder (vocab=32000, d=416) &  -- / -- / -- & 13,312,000 \\
BlockDecoder (atten h=32, key value h=32, head d =52)  & 1664 / 4096 / 8  & 252,210,816 \\
 Ctx Converter (in d = 1664, out d = 1664) & -- / -- / -- & 5,541,120 \\
 Embedder (vocab=32000, d=1664) &  -- / -- / -- & 53,248,000 \\
TokenDecoder (atten h=32, key value h=32, head d =52)   & 1664 / 4096 / 8  & 252,210,816 \\
   LM Head                               & 1664 / -- / -- & 53,248,000  \\
\midrule
Total                                          & --       & 629,770,752 \\
\bottomrule
\end{tabular}
\end{table*}
\begin{table*}[tb]
\centering
\caption{Parameter breakdown for the Block Transformer (900M).}
\label{tab:pb900M}
\begin{tabular}{lrrr}
\toprule
Module & Hidden / Int. / Layers & Params  \\
\midrule
Embedder (vocab=32000, d=448) &  -- / -- / -- & 14,336,000 \\
BlockDecoder (atten h=32, key value h=32, head d =56)  & 1792/4608/10  & 376,214,272 \\
 Ctx Converter (in d = 1792, out d = 1792) & -- / -- / -- &  6,426,112 \\
 Embedder (vocab=32000, d=1792) &  -- / -- / -- & 57,344,000 \\
TokenDecoder (atten h=32, key value h=32, head d =56)   & 1792/4608/10  & 376,214,272 \\
   LM Head                               & 1792 / -- / -- & 57,344,000  \\
\midrule
Total                                          & --       & 887,878,656 \\
\bottomrule
\end{tabular}
\end{table*}
\begin{table*}[tb]
\centering
\caption{Parameter breakdown for the Block Transformer (1.2B).}
\label{tab:pb1.2B}
\begin{tabular}{lrrr}
\toprule
Module & Hidden / Int. / Layers & Params  \\
\midrule
Embedder (vocab=32000, d=480) &  -- / -- / -- & 15,360,000 \\
BlockDecoder (atten h=32, key value h=32, head d =60)  & 1920 / 5120 / 12  & 530,889,600 \\
 Ctx Converter (in d = 2048, out d = 2048) & -- / -- / -- & 7,376,640 \\
 Embedder (vocab=32000, d=1664) &  -- / -- / -- & 61,440,000 \\
TokenDecoder (atten h=32, key value h=32, head d =60)   & 1920 / 5120 / 12  & 530,889,600 \\
   LM Head                               & 1920/ -- / -- & 61,440,000  \\
\midrule
Total                                          & --       & 1,207,395,840 \\
\bottomrule
\end{tabular}
\end{table*}
\begin{table*}[tb]
\centering
\caption{PHOTON (600M)}
\label{tab:pp600M}
\begin{tabular}{llrrr}
\toprule
Level & Module & Hidden / Int. / Layers & Params \\
\midrule
& Embedder (vocab=32000, d=416) &  -- / -- / -- & 13,312,000 \\
\multirow{2}{*}{Enc. ($l=1$)}
  & Ctx Chunker (block=4, concatenate) & -- / -- / -- & -- \\
  & Ctx Encoder (atten h=32, kv h=32, head d=52) & 1664 / 4096 / 4 & 126,106,240 \\
\multirow{2}{*}{Enc. ($l=2$)}
  & Ctx Chunker (block=4, linear) & -- / -- / -- & 11,083,904 \\
  & Ctx Encoder (atten h=32, kv h=32, head d=52) & 1664 / 4096 / 4 & 126,106,240 \\
\multirow{2}{*}{Dec. ($l=2$)}
  & Ctx Converter (in d=1664, out d=1664) & -- / -- / -- & 5,541,120 \\
  
  & Ctx Decoder (atten h=32, kv h=32, head d=52) & 1664 / 4096 / 4 & 126,106,240 \\
\multirow{4}{*}{Dec. ($l=1$)}
  & Ctx Converter (in d = 1664, out d = 1664) & -- / -- / -- & 5,541,120 \\
& Embedder (vocab=32000, d=1664) &  -- / -- / -- & 53,248,000 \\
  & Ctx Decoder (atten h=32, kv h=32, head d=52) & 1664 / 4096 / 4 & 126,106,240 \\
  &  LM Head (in d=1664, out d=32000)                               & 1664 / -- / -- & 53,248,000  \\
\midrule
Total & -- & -- & 646,399,104 \\
\bottomrule
\end{tabular}
\end{table*}

\begin{table*}[tb]
\centering
\caption{PHOTON (900M)}
\label{tab:pp900M}
\begin{tabular}{llrrr}
\toprule
Level & Module & Hidden / Int. / Layers & Params \\
\midrule
& Embedder (vocab=32000, d=448) &  -- / -- / -- & 14,336,000 \\
\multirow{2}{*}{Enc. ($l=1$)}
  & Ctx Chunker (block=4, concatenate) & -- / -- / -- & -- \\
  & Ctx Encoder (atten h=32, kv h=32, head d=56) & 1792 / 4608 / 5 & 188,108,032 \\
\multirow{2}{*}{Enc. ($l=2$)}
  & Ctx Chunker (block=4, linear) & -- / -- / -- & 12,854,016 \\
  & Ctx Encoder (atten h=32, kv h=32, head d=56) & 1792 / 4608 / 5 & 188,108,032 \\
\multirow{2}{*}{Dec. ($l=2$)}
  & Ctx Converter (in d=1792, out d=1792) & -- / -- / -- & 6,426,112 \\
  
  & Ctx Decoder (atten h=32, kv h=32, head d=56) & 1792 / 4608 / 5 & 188,108,032 \\
\multirow{4}{*}{Dec. ($l=1$)}
  & Ctx Converter (in d = 1792, out d = 1792) & -- / -- / -- & 6,426,112 \\
& Embedder (vocab=32000, d=1792) &  -- / -- / -- & 57,344,000 \\
  & Ctx Decoder (atten h=32, kv h=32, head d=56) & 1792 / 4608 / 5 & 188,108,032 \\
  &  LM Head (in d=1792, out d=32000)                               & 1792 / -- / -- & 57,344,000  \\
\midrule
Total & -- & -- & 907,162,368 \\
\bottomrule
\end{tabular}
\end{table*}

\begin{table*}[tb]
\centering
\caption{PHOTON (1.2B)}
\label{tab:pp1.2B}
\begin{tabular}{llrrr}
\toprule
Level & Module & Hidden / Int. / Layers & Params \\
\midrule
& Embedder (vocab=32000, d=480) &  -- / -- / -- & 15,360,000 \\
\multirow{2}{*}{Enc. ($l=1$)}
  & Ctx Chunker (block=4, concatenate) & -- / -- / -- & -- \\
  & Ctx Encoder (atten h=32, kv h=32, head d=60) & 1920 / 5120 / 6 & 265,445,760 \\
\multirow{2}{*}{Enc. ($l=2$)}
  & Ctx Chunker (block=4, linear) & -- / -- / -- & 14,755,200 \\
  & Ctx Encoder (atten h=32, kv h=32, head d=60) & 1920 / 5120 / 6 & 265,445,760 \\
\multirow{2}{*}{Dec. ($l=2$)}
  & Ctx Converter (in d=9728, out d=2432) & -- / -- / -- & 7,376,640 \\
  & Ctx Decoder (atten h=32, kv h=32, head d=60) & 1920 / 5120 / 6 & 265,445,760 \\
\multirow{4}{*}{Dec. ($l=1$)}
  & Ctx Converter (in d = 2432, out d = 2432) & -- / -- / -- & 7,376,640 \\
  & Embedder (vocab=32000, d=1920) &  -- / -- / -- & 61,440,000 \\
  & Ctx Decoder (atten h=32, kv h=32, head d=60) & 1920 / 5120 / 6 & 265,445,760 \\
    &  LM Head (in d=1920, out d=32000)                               & 1920 / -- / -- & 61,440,000  \\
\midrule
Total & -- & -- & 1,229,531,520 \\
\bottomrule
\end{tabular}
\end{table*}

\end{document}